\documentclass[reqno,12pt]{amsart}
\usepackage{times}
\usepackage{graphicx}
\usepackage{hyperref}
\usepackage{lipsum}
\usepackage[normalem]{ulem}
\usepackage[hang,small,bf]{caption}
\usepackage[T2A]{fontenc}
\usepackage[utf8]{inputenc}
\usepackage[backend=bibtex,bibencoding=utf8,citestyle=authoryear]{biblatex}
\newcommand{\E}{\mathbb{E}}
\newcommand{\V}{\mathbb{V}}

\begin{document}
\title{ Variation of word frequencies in Russian literary texts}
\thanks{e-mail: vladislav.kargin@gmail.com}
\date{May 2015}
\author{Vladislav Kargin}
\maketitle

\begin{center}
\textbf{Abstract}
\begin{quotation}
We study the variation of word frequencies in Russian literary texts. Our findings indicate that the standard deviation of a word's frequency across texts depends on its average frequency according to a power law with exponent $0.62,$ showing that the rarer words have a relatively larger degree of frequency volatility (i.e., ``burstiness'').   
 
  Several latent factors models have been estimated to investigate the structure of the word frequency distribution. The dependence of a word's frequency volatility on its average frequency can be explained by the asymmetry in the distribution of latent factors.       
\end{quotation}
\end{center}

\bigskip

\section{Introduction}

The study of word frequency variation in different texts arose first in the problem of author attribution ( \cite{zipf32}, \cite{yule44}, \cite{mosteller_wallace64}). Recently, the explosive growth in the computing power and in the text data volume led to many new applications.  For example, the text indexing problem asks to associate documents with queries for fast retrieval; the authorship profiling problem require to describe features of the author (sex, age, religious and political beliefs, etc) based on texts that the author produced. In addition, the classic authorship attribution problem found new applications in security and forensics 
(see surveys by \cite{holmes98}, \cite{juola08}, \cite{koppel_schler_argamon09} and \cite{stamatatos09}). 

For all these applications, the fundamental statistical issue is the distribution of word frequencies\footnote{In this paper we use the term ``frequency'' as usual in statistics, that is, the number of the word occurrences in a document divided by the document's total number of words.}  in different texts. For example, if a word in a query has its frequency in a document higher than its average frequency, then this document can be regarded as more relevant to the query. 

Some properties of the word frequency distribution were noticed a long time ago. For example, Zipf's law (\cite{zipf32}) describes the distribution of word frequencies in a particular text, and Heaps' law (p. 207 in \cite{heaps78}, p.75 in \cite{herdan66}) relates the number of distinct words in a text to its length. Some new research on these laws was done in  \cite{fbc13}, \cite{gerlach_altman13}, \cite{gerlach_altman14}, and \cite{piantadosi14}. See also surveys in \cite{zanette14}, \cite{gerlach_altman15}. This paper focuses on a different set of properties and investigates the variation of word frequencies across documents.

  One has to understand the structure of the word-document frequency matrix for applications in the information retrieval, in order to handle the problems of word synonymity and polysemy. For this purpose, there have been recently developed tools such as LSA (``latent semantic analysis'', \cite{ddflh90}), pLSA (``probabilistic latent semantic analysis'', \cite{hofmann99}), and LDA (``latent Dirichlet allocation'', \cite{blei_ng_jordan03}).  The main idea of  these methods is the dimension reduction. The variation of word frequencies across texts is assumed to stem mainly from the variation in relatively small amount of factors (or ``topics'') across texts.   

The goal of this study is to establish basic facts about the fluctuations of word frequencies across documents such as the dependence of the fluctuation size on the average word frequency. In order to clarify this dependence, we will apply the latent factor techniques such as LSA, pLSA, and LDA.

The data for this study come from a large online library of Russian literary texts. This collection is especially appropriate for our study since it covers a very large spectrum of texts from various authors, genres and epochs.   

The paper is organized as follows. First, in Section \ref{section_data} we describe the data. Then, in Section \ref{section_variation} we study how the size of frequency fluctuations across texts depends on the word's average frequency. Next, in Sections \ref{section_LSA} and \ref{section_LDA} we apply factor models to analyze the variation of vocabulary across texts in more detail. Finally, Section \ref{section_conclusion} concludes.

\section{A preliminary look at the data}
\label{section_data}
We use data from Flibusta, a Russian online library. It covers Russian and translated fiction works from many historical periods and literary genres.
Currently, it has between $200,000$ and $300,000$ texts by about $85,000$ authors, where the author is
understood to include translators and sometimes organizations that published
a particular text. Our analysis uses only a part of this dataset (around $25,000$ books). In particular, we use only books which are available in a text format (more precisely, in the ``FB2'' book format) and we exclude the documents that are available only as pdf, djvu, doc, and other binary formats.

The library works using the wiki principle and the texts
are uploaded by users, therefore the number of texts depends both on how many texts were written by the author and on how many of them were uploaded by users. Table \ref{top_authors} in Appendix shows authors with the largest number of texts.

The top place belongs to ``Unknown Author'', which can be associated with texts such
a ``Bhagavad Gita'' or ``Poetry of Medieval France''. In the second place one sees a weekly political publication
"Tomorrow". 

The third and fourth places belong to the American and Russian science fiction writers Ray Bradbury and Kir Bulychev, respectively.  Many of the other top authors are authors and translators of books in popular genres such as science fiction, mystery, romance, action,  historical fiction, sensational and how-to literature.

The right portion of Table \ref{top_authors} in Appendix shows the top $25$ authors after we excluded the ``unknown author'', weekly publications, translators, and the authors working in the genres associated with popular culture. The result is the list of well-known authors, most of which are short story writers. For these authors, the number of texts in the online library ranges from $446$ for Anton Chekhov to $144$ for Franz Kafka.

\section{Variation of word frequencies across texts}
\label{section_variation}

 Suppose that $\xi_{b,w}^{(t)}$ is an indicator variable which equals $1$ if the word at place $t$ in book $b$ equals $w$. Then, the frequency of word $w$ in book $b$ can be written as 
\begin{equation}
x_{b,w}=\frac{1}{T_b}\sum_{t=1}^{T_b}{\xi_{b,w}^{(t)}},
\end{equation}
where $T_b$ is the length of the book $b$.

First, let us take the hypothesis that for a given $w$ the variables $\xi_{b,w}^{(t)}$ are  independent identically distributed random variables with the expectation parameter $p_w$, which does not depend on $b$.  Then $\E x_{b,w} = p_w$, and 
\begin{equation}
\V \left( x_{b,w} \right)= \frac{p_w(1-p_w) }{T_b}  .
\label{variance_w}
\end{equation}

\begin{table}[htbp]
  \begin{tabular}{c}
    \hline \\
		  \includegraphics[width=.5\textwidth]{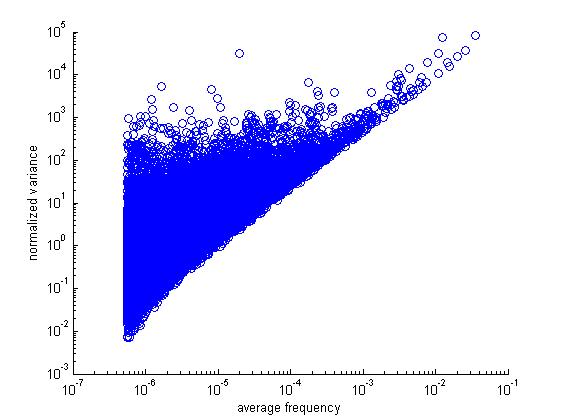}
	 \\ 
    \hline
  \end{tabular}%
\captionsetup{name=Figure}
\caption{ Normalized variance vs average frequency. }
\label{fig_norm_var}
\end{table}

We can use (\ref{variance_w}) to check $\xi_{b,w}^{(t)}$ are i.i.d. variables. For this purpose, we estimate $p_w$ by using the whole sample: 
\begin{equation}
\widehat{p}_{w}:=\frac{1}{T}\sum_{b=1}^{B}\sum_{t=1}^{T_b}{\xi_{b,w}^{(t)}},
\end{equation}
 where $T$ is the total number of words in the data and $B$ is the number of texts, and then we compute the normalized variance of $x_{b,w}$ across books.
\begin{equation}
V_w=\frac{1}{B}\sum_{b=1}^{B} \left( \frac{\sqrt{T_w}(x_{b,w}-\widehat{p}_{w})}{\sqrt{\widehat{p}_{w}(1-\widehat{p}_{w})}} \right)^2.
\label{empricial_variance_w} 
\end{equation}
 This statistic should be compared with $1$.

The results are shown in Figure \ref{fig_norm_var}. They suggest that this model is not acceptable and that there is a significant degree of variation in the distribution of $\xi_{b,w}$ across texts. 

This variation in the word frequency distribution across texts is at the heart of most applications. However, its first systematic study is relatively recent and was done in \cite{church_gale95}. The phenomenon is often called \textit{burstiness} for a measure of word frequency variability which was used in Church and Gale.\footnote{The name ``burstiness'' comes from the observation that if a rare word has occurred at least once in a document, then it is likely to occur more times in the same document than  it is predicted by a Poisson distribution with the word's average frequency. This observation can be explained by the variability of the word frequencies across texts since the observation of a word in a document changes the posterior belief about the word frequency in this document.}   One interesting observation of Church and Gale is that the words with an unusually high frequency variability are often content words: they have an additional linguistic load.

Now, let the variables $\xi_{b,w}^{(t)}$ be independent random variables, which are identically distributed conditional on $b$ and $w$ and have the expectation parameter $p_{b,w}$. That is, the parameter is allowed to change from text to text and we are interested in learning how it is distributed across texts.

 The simplest estimate for $p_{b,w}$ is $x_{b,w}=\frac{1}{T_b}\sum_{t=1}^{T_b}{\xi_{b,w}^{(t)}}$.
It is reliable only if the standard deviation of the estimate is sufficiently small:
\begin{equation}
p_{b,w}\gg \sqrt{\frac{p_{b,w}(1-p_{b,w})}{T_b}},
\end{equation}
or $p_{b,w} \gg T_b^{-1}$.

In our database, the average text length is of the order of $3\times 10^4$ words and therefore we can expect that $x_{b,w}$ reliably 
estimates $p_{b,w}$ only if $p_{b,w} \geq 10^{-4}$.

Let us define the average word frequency:
\begin{equation}
\overline{x}_{w}=\frac{1}{B}\sum_{b=1}^{B}x_{b,w},
\end{equation} 
and the cross-text variance: 
\begin{equation}
\sigma^2_{w}=\frac{1}{B}\sum_{b=1}^{B}(x_{b,w}-\overline{x}_{w})^2.
\end{equation}

In the next pictures we order word types by their average frequency. 

\begin{table}[htbp]
  \begin{tabular}{cc}
    \hline \\
 		\begin{minipage}[t]{0.5\textwidth} 
		  \includegraphics[width=\textwidth]{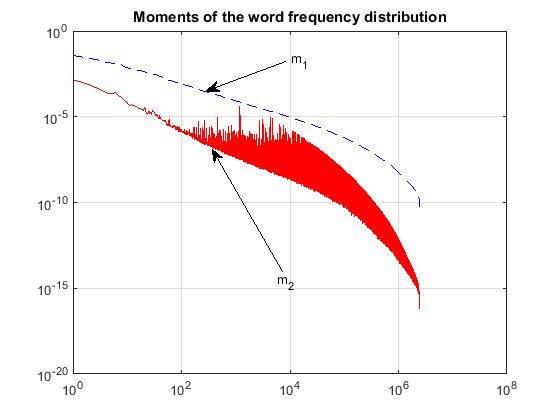} 
		\end{minipage} 
		& 
		\begin{minipage}[t]{0.5\textwidth} 
		  \includegraphics[width=1\textwidth]{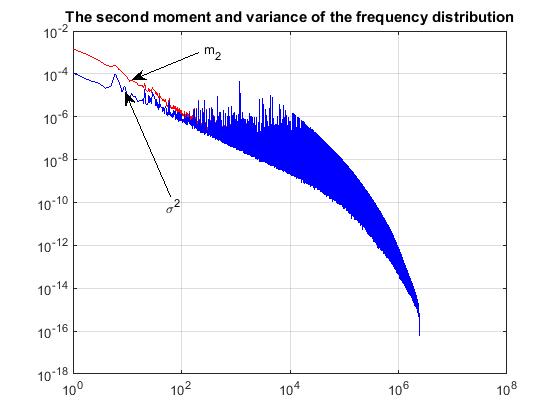}
    \end{minipage}
     \\ 
    \hline
  \end{tabular}%
\captionsetup{name=Figure}
\caption{ The (estimated) expectation, second moment, and variance of the word frequency distribution.}
\label{fig_log_moments}
\end{table}

The pictures in Figure \ref{fig_log_moments} suggest that in general the variance decline together with the average frequency, soe it is natural to ask about the law of this dependence. 

\begin{table}[htbp]
  \begin{tabular}{c}
    \hline \\
		  \includegraphics[width=.5\textwidth]{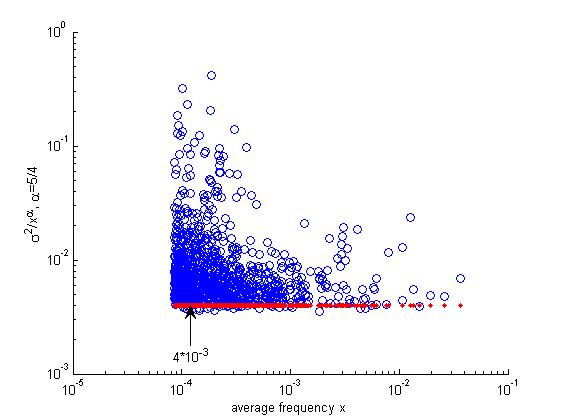}
	 \\ 
    \hline
  \end{tabular}%
\captionsetup{name=Figure}
\caption{ Normalized variance vs average frequency; $1000$ of the most frequent words. }
\label{fig_norm_var2}
\end{table}

In Figure \ref{fig_norm_var2}, the vertical axis shows the variance normalized by a power of the average frequency:
\begin{equation}
y_w=\frac{\sigma_w^2}{\overline{x}_w^{1.25}}.
\end{equation} 
The exponent $\kappa=1.25$ was chosen to fit the data. 
We show the results for $1,000$ words with the largest frequency. These are the words for which we can expect
 that the variance $\sigma_w^2$ is reliably estimated. Figure \ref{fig_norm_var2} demonstrates that the variance follows the power law:  

\begin{equation}
\sigma^2\sim a \overline{x}^{1.25},
\end{equation}
where $a$ is a random variable which generally exceeds $4 \times 10^{-3}.$ 

Or, in terms of the ratio of the standard deviation to the mean: 

\begin{equation}
\frac{\sigma}{\overline{x}}\sim a^{1/2} \overline{x}^{-0.375}.
\end{equation}

That is, the ratio increases for rarer words.

\begin{table}[htbp]
  \begin{tabular}{c}
    \hline \\
		  \includegraphics[width=.5\textwidth]{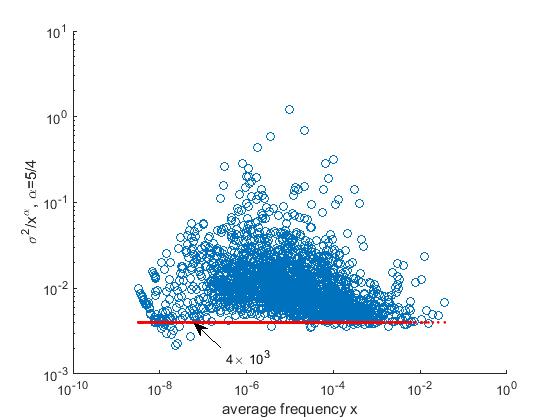}
	 \\ 
    \hline
  \end{tabular}%
\captionsetup{name=Figure}
\caption{ Normalized variance vs average frequency; a sample of $2000$ words.}
\label{fig_norm_var2_2000}
\end{table}

Figure \ref{fig_norm_var2_2000} is similar except it also shows variances and average frequencies for some of the less-frequent words. (We simply use the first $2000$ different words that appeared in the data.) The conclusion drawn from Figure \ref{fig_norm_var2} is not changed by Figure \ref{fig_norm_var2_2000}, although we observe some deviations below the straight line for the normalized variance of less-frequent words. 

In summary, these observations show that there is a power dependence between the variance of document word frequencies and the average frequency. The frequent words have larger variation in frequency across texts. However, the ratio of the standard deviation to the average frequency is increases as the average frequency becomes smaller.  This dependence follows a power law with an exponent of approximately $-0.375$.

This relation can be seen as a quantification of the burstiness phenomenon. In particular, it shows that burstiness is in general more pronounced for rarer words. Hence, if volatility of a word's frequency (i.e., its burstiness) is used to evaluate the amount of content associated with the word, then the volatility should be normalized by a function of its frequency.   

 In the next section, we will try to uncover the structure in the variation of document word frequencies using a factor model, which is a variant of the LSA model. 

\section{A factor model for the vocabulary size variation}
\label{section_LSA}

In a factor model, expected word frequencies are allowed to change from text to text, as in the general random effects model. However, it is postulated that these changes can be explained by a relatively small number of factors. This approach is especially convenient for very large collections of data, when we are interested in reducing the complexity of the data, or, in other words, in ``reducing the dimensionality'' of an observed phenomenon.

 Let the empirical frequency distribution of word types in a book $b$ is
denoted $x_{b}$. If the number of word types
is $N,$ then $x_{b}$ is an $N$-vector. Each entry $(x_b)_w$ is the frequency of word type $w$ in book $b$. In particular, 
 $\left\Vert x_{b}\right\Vert _{1}=1.$ 

The simplest factor model, which is a variant of the LSA model, assumes that $x_b$ has a part which can be explained by a small number of factors and a part which is an unexplained noise. 
Hence the model is  
\begin{equation}
X=\sum_{k=1}^{s} \theta_{k} f_k  v_k^{\ast} + Z,
\label{model_finite}
\end{equation}%
where $X$ is an $N$-by-$B$ matrix whose columns are $x_b$, the empirical frequency distributions of word
types, and where $Z$ is a noise matrix.\footnote{The difference from the original LSA model is that here the decomposition is applied to the frequency matrix $X$ rather than to the matrix of word counts in each document. In addition, the more recent implementations of the LSA method usually use ``tf-idf'' (term frequency, inverse document frequency) instead of raw counts.  This correction often improves performance of the LSA in document indexing tasks. We will not use this modification in our version since it essentially removes the frequent words (like ``the'' and ``in'') from consideration, and these words were found important in other tasks such as authorship attribution.}  
We assume that $\{f_k\}$ is an orthonormal system of $N$-vectors, and $\{v_k\}$ is an orthonormal system
of $B$-vectors. 

Every book $b$ can be characterized by vector $\omega_b=((v_1)_b, \ldots, (v_s)_b)$, and books with 
the same vector $\omega$ are expected to have the same word frequency distribution up to noise. 

 The simplest method is to estimate  $\theta_k$,  $f_k$, and $v_k$ is by computing the SVD (``Singular Value Decomposition'') of the matrix $X$ and to use only that part of the decomposition that corresponds to large singular values. 

There are several benefits of this model. First, it has a straightforward interpretation: the frequency matrix is approximated by a small-rank matrix. Hence, we fit a parsimonious model to the data and have a clear trade-off between the quality of the approximation and the complexity of the model. Second, the model can be estimated with efficient and fast SVD algorithms. Finally, the statistical literature about factor models is rich and may provide some guidance about the choice of the number of factors.

There are also significant deficiencies. The most important is that the model ignores the fact that $x_b$ are the empirical frequency distributions. This is especially troublesome for less-frequent words, when most of the entries in $x_b$ are zeros.  

The second deficiency is that most of the results about the number of factors are derived under the assumption that $Z$  has i.i.d Gaussian entries. In our situation, this assumption does not hold.

From the computational prospective, matrix $X$ is very large (of order $10^4$ by $10^6$), and it is computationally difficult to estimate its spectral parameters. There are two alternative approaches to handle this difficulty. 

First, one can take a sample of texts and analyze the spectral data using this sample. Second, the text-word matrix can be restricted to the part that contain only the most frequent word types. 

In this paper, we choose the second method that uses the most frequent words.

In particular, we computed eigenvalues $\theta_k$ and eigenvectors $f_k$ for $500$ most frequent words. 
For the applications, one also need to know $v_k$, the eigenvectors of a large $B$-by-$B$ matrix $X^{\ast} X$. Fortunately, they can be easily computed: 
\begin{equation}
v_k=\frac{1}{\theta_k}X^{\ast} f_k.
\end{equation}

The four largest eigenvalues were found equal to : $91.6,$ 
   $3.15,$ $1.76,$ and $1.52.$  The first eigenvalue is much larger than the other ones and corresponds to an eigenvector with positive entries. This eigenvector can be interpreted as the average frequency distribution and all other eigenvectors as ``corrections''. 
	
	In order to estimate the number of factors, we note some stylized facts from the theory of large random matrices
	(\cite{baik_silverstein06}, \cite{paul07}, \cite{BenaychGeorges_Nadakuditi12}). If a large random matrix $Z$ deformed by a low-rank matrix, then the resulting matrix $X$ has the ``bulk'' spectrum that correspond to singular values of $Z$ and outlier singular values which correspond to the singular values of the low-rank perturbations.

 The plots for the eigenvalues and their spacings suggest that there are  at least $10$ outliers that can be interpreted as detectable factors. The plot of eigenvectors $f_k$ suggest that the eigevectors are concentrated on less than 100  of the most frequent words.

\begin{table}[htbp]
  \begin{tabular}{cc}
    \hline \\
    \begin{minipage}[t]{0.45\textwidth} 
		  \includegraphics[width=1\textwidth]{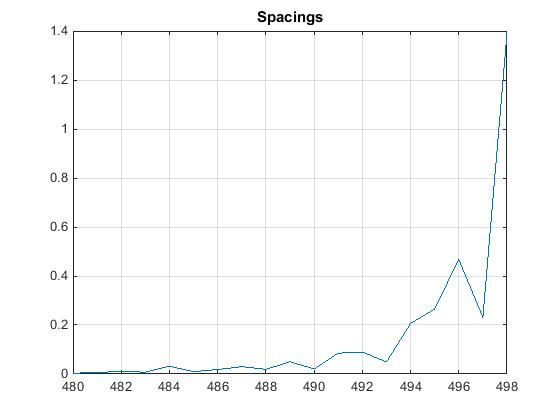}
    \end{minipage} 
		& 
		\begin{minipage}[t]{0.46\textwidth} 
		  \includegraphics[width=\textwidth]{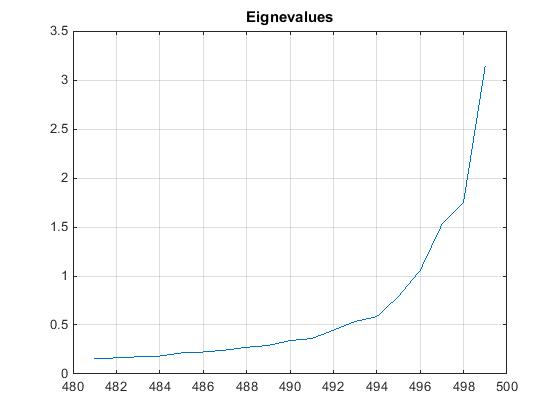} 
		\end{minipage} \\ 
    \hline
  \end{tabular}%
\captionsetup{name=Figure}
\caption{ Distribution of eigenvalues of $X X^{\ast}$. (The largest eigenvalue is excluded.)}
\label{fig_eig_500_words_smaller}
\end{table}

	\begin{table}[htbp]
  \begin{tabular}{c}
    \hline \\
    \begin{minipage}[t]{\textwidth} 
		  \includegraphics[width=.7\textwidth]{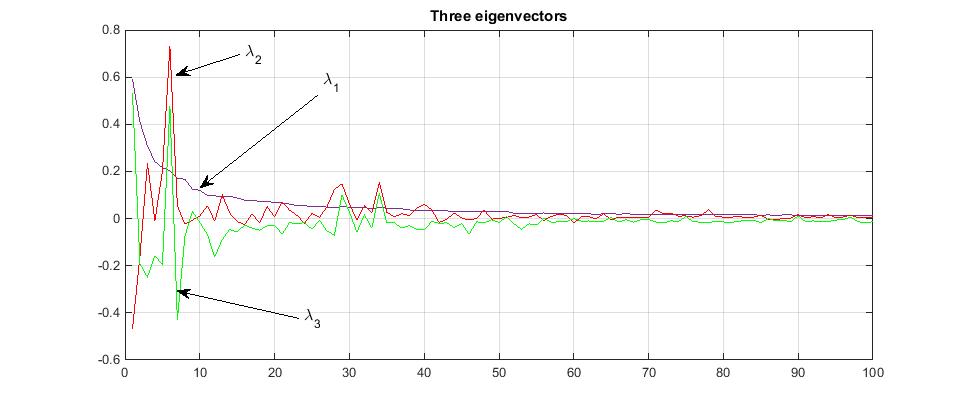}
    \end{minipage} 
  \end{tabular}%
\captionsetup{name=Figure}
\caption{ The eigenvectors $f_k$ for the first three largest eigenvalues of $X X^{\ast}$. }
\label{fig_evectors_500_words}
\end{table}

\section{pLSA and LDA models}
\label{section_LDA}

In the pLSA (``probabilistic latent semantic analysis'') approach, the true word frequencies in a document are modeled as a mixture of a few probability distributions, which are interpreted as word distributions belonging to a factor (or a ``topic'' in the terminology of text indexing literature). 
\begin{equation}
P(w|b)=\sum_{z=1}^{s}P(w|z)P(z|b)
\label{model_pLSI}
\end{equation}
The interpretation is that for each word in a book $b$ we randomly select a topic $z$ and then select the probability of a word $w$ on the basis of this topic.  In other words, given topic $z$, the probability of a word $w$ is independent of  the book $b$.
The model  resembles the factor model (\ref{model_finite}). However, its strong advantage is that this model treats the word frequencies as a probability distribution in a true probability model.

Assuming further the independence of word frequencies in a document, the model can be estimated by the log-likelihood maximization with the following log-likelihood function:
\begin{equation}
\mathcal{L}=\sum_{b,w} n_{w,b}\log P(w|b),
\end{equation}
where $n_{b,w}$ is the number of occurences of the word $w$ in a book $b$.
The maximization can be performed by the EM method, although as usual, there is a problem of local maxima. In addition, in our experience the convergence rate was rather slow.\footnote{ In the model with 10 factors, 100 frequent words and approximately 27,000 books, the convergence from a random starting guess to 
the 6th digit took several minutes and to the 9th digit took several hours, with some evidence that each new digit of precision takes progressively more time. The code was implemented in Matlab on a PC machine. We have also used the pLSA code from \cite{verbeek06} for comparison. It yielded similar results.}

The  LDA (``latent Dirichlet allocation'') model is similar to the pLSA in that it is assumed that the distribution of words in a text is controlled by an $s$-by-$N$ matrix $\beta$ which is a matrix of conditional probability of a word given a topic, 
 $\beta_{zw}=P(w|z)$. Every document is associated with a probability distribution over topics $\theta_b$ which is an $s$-vector of conditional probabilities $ (\theta_b)_z=P(z|b)$. 
The novel idea is to treat the vector $\theta_b$ as a random variable drawn from a Dirichlet distribution with an $s$-vector parameter $\alpha$.
 
The idea to treat conditional probabilities as random variables is the key idea of the hierarchical Bayesian modeling. In this particular context, its main intention is to use the information about the distribution of $\theta_b$ over all texts $b$ in order to make more precise estimates of a particular $\theta_b$.

To restate, the joint distribution of the mixture $\theta$, and sequences of words $\{w_i\}$  and topics $\{z_i\}$ in a text $b$ is 
\begin{equation}
P(\theta, \{w_i\}, \{z_i\}|\alpha,\beta)=P(\theta|\alpha) \sum_{i=1}^{N_b}P(w_i|z_i,\beta)P(z_i|\theta),
\label{model_LDA}
\end{equation}
where $P(\theta|\alpha)$ is the Dirichlet distribution with parameter $\alpha$.

The main task is to estimate the parameters $\alpha$ and $\beta$ and compute the posterior distribution $P(\theta|\{w_i\})$. This is a non-trivial computational problem. Several approximation algorithms are available. For details, see paper  by \cite{blei_ng_jordan03}. In our experiments we used the code developed in \cite{verbeek06}. 

The evaluations of practical benefits of LDA over pLSA differ. While \cite{blei_ng_jordan03} found some benefits of the LDA over pLSA in the context of collaborative filtering, \cite{masada08} found no advantage of LDA over pLSA in classification of Japanese and Korean webpages. 

The advantage of the LDA model for our purposes is that it can be used to investigate the burstiness phenomenon. (For a related model, the Dirichlet compound multinomial model, the burstiness was investigated in \cite{madsen05}.)

In particular, we will use the LDA model to clarify results found in Section \ref{section_variation}. First, note that the probability of word $w$ in a book $b$ equals $(\theta \beta)_{bz} = \sum_{z=1}^s \theta_{bz} \beta_{zw}.$ Here $\theta_b$ is a realization of a random vector $\theta$ distributed according to the Dirichlet distribution with parameter $\alpha$. The joint moments of the Dirichlet distribution are well-known: 
\begin{equation*}
E\left(\prod_{z=1}^s \theta_i^{k_i}\right ) = \frac{\Gamma\left(\sum_i{\alpha_i}\right)}{\Gamma\left( \sum_i (\alpha_i+k_i)\right)}
\times \prod_i  \frac{\Gamma\left(\alpha_i + k_i\right)}{\Gamma\left(  \alpha_i\right)},
\end{equation*}
and therefore one can easily compute the moments of the linear combinations of $\theta_i$. 

Consider, for simplicity, the case with only two factors and the symmetric Dirichlet distribution. So, let $s=2$ and $\alpha_1=\alpha_2=\alpha$. Then the probability that a particular word in a book is a word $w$ has a distribution with the expectation:
\begin{equation*}
\E(p_w)=\frac{1}{2}(\beta_{1w}+\beta_{2w})
\end{equation*}
and the variance can be computed as 
\begin{equation*}
\V(p_w)=\frac{1}{4}\frac{1}{2\alpha +1}(\beta_{1w}-\beta_{2w})^2.
\end{equation*}

If $\xi_w=|\beta_{2w}-\beta_{1w}|/2$, then we could recover the findings in Section \ref{section_variation} provided that 
$\xi_w \sim  (\E p_w)^{\kappa/2}$ with $\kappa=1.25$. The problem with this interpretation, is that this relation is impossible for small $\E p_w$. Indeed, the positivity of $\beta_{1w}$ and $\beta_{2w}$ implies that $\xi_w \leq \E p_w$ and this contradicts the previous relation for small $\E p_w$.  This can also be seen from the fact that $\V(p_w) \leq (\E p_w)^2$ in this model. 
 
This can be rectified by using an asymmetric model. Take for example $s=2$, $\alpha_1=1$ and $\alpha_2=\alpha$. In this case, 
\begin{equation*}
\E(p_w)=\frac{1}{1+\alpha}\beta_{1w}+\frac{\alpha}{1+\alpha}\beta_{2w},
\end{equation*}

\begin{equation*}
\V(p_w)=\frac{\alpha}{(2+\alpha)(1+\alpha)^2}(\beta_{1w}-\beta_{2w})^2.
\end{equation*}

Let $\alpha \ll 1$, $\beta_{1w}=\gamma_w \alpha \ll \beta_{2w} $. Then, 
\begin{equation*}
\left[\E(p_w)\right]^2 \sim (\gamma_w + \beta_{2w})^2 \alpha^2,
\end{equation*}

and 
\begin{equation*}
\V(p_w) \sim \frac{\beta_{2w}^2}{2} \alpha.
\end{equation*}

Hence,  $\V(p_w) \gg \left[\E(p_w)\right]^2$ provided that $\gamma_w$ is not too large relative to $ \beta_{2w}.$

Intuitively, the second topic occurs very rarely ($\alpha \ll 1$). However, it is associated with much larger conditional probability to observe the word $w$: $\beta_{2w} \gg \beta_{1w}$. This leads to a relatively large variance of the frequency distribution for the word $w$. In other words, the high burstiness of the word $w$ is due to its being a marker of a rare topic.  

Next, we observe that when $\alpha$ is small and fixed, the power relation $\V(p_w)=\left[\E(p_w)\right]^{1.25}$ is possible but only if 
$\gamma_w \gg \beta_{2w}$. Since $\gamma_w=\beta_{1w}/\alpha$, it follows that the relationship can occur in a limited range when 
$\beta_{1w} \ll \beta_{2w} \ll \beta_{1w}/\alpha.$  This range is wide only if $\alpha$ is small

\begin{table}[htbp]
  \begin{tabular}{cc}
    \hline \\
 		\begin{minipage}[t]{0.5\textwidth} 
		  \includegraphics[width=\textwidth]{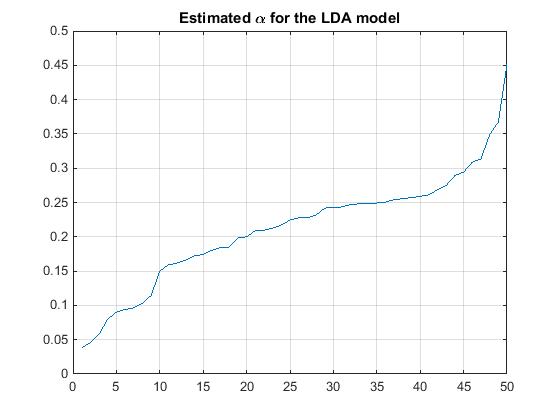} 
		\end{minipage} 
		& 
		\begin{minipage}[t]{0.5\textwidth} 
		  \includegraphics[width=1\textwidth]{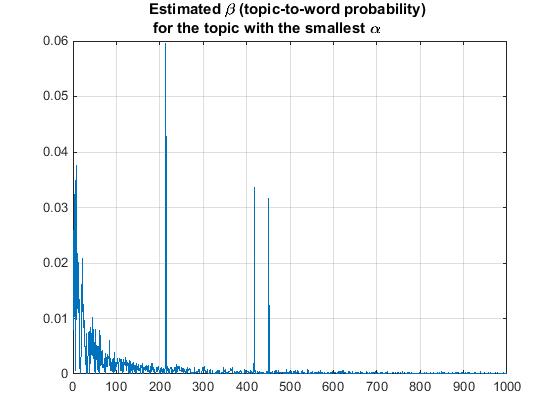}
    \end{minipage}
     \\ 
    \hline
  \end{tabular}%
\captionsetup{name=Figure}
\caption{ The estimated parameters of the LDA model with $50$ topics and $1000$ most frequent words.}
\label{fig_LDA_estimates}
\end{table}

In summary, the power relation observed in Section 3 appears to be due to the asymmetry in the distribution of topics vector $\theta$, and, in particular, it is due to the existence of rare topics that are associated with some specific words (``topic markers'').

In order to demonstrate the asymmetry in the distribution of topics in the data, we show the estimates of the parameter $\alpha$ which is the Dirichlet parameter for topics, and the parameter $\beta_z=p(w|z)$ for one of the rare topics $z$.

 The left plot in Figure \ref{fig_LDA_estimates} shows the distribution of $\alpha$, which ranges from $0.04$ to $0.45$. The right plot shows that a rare topic is indeed associated with marker words. In this example, for the topic with $\alpha=0.04$, there are three relatively infrequent words with $\beta>0.03$. They are ``всё'' (``all''), ``ещё'' (``yet''), and ``её'' (``her''). Their average frequencies are 
$3.8\times 10^{-4}$, $2\times 10^{-4}$, and $1.9\times 10^{-4}$, respectively. The common feature of these words is the presence of 
the letter ``ё''. This letter is often substituted by the letter ``е'' to economize on typography costs, and its presence indicates that either the book is intended for children or it has been published recently with the help of computerized typography.

\section{Conclusion}
\label{section_conclusion}
In this paper we studied the variation in the vocabulary of Russian literary texts from a large online database.  

First, we detected a significant variation in the distribution of word frequencies across texts, and found that the variance of this distribution is in general larger for words with higher frequency. We found that the dependence of the word frequency volatility  on its mean has a form of power law with the exponent
 $0.625,$ which quantify the observation that rarer words has greater degree of ``burstiness''.

 In order to study the variation in word frequencies across texts, we applied several variants of the factor analysis method.  We found that most of the variation is concentrated in approximately 100 functional words and a significant portion of this variation can be explained by about 10 factors. An analysis of the LDA model suggests that the power dependence of the frequency volatility on its mean can be explained by an asymmetry in the prior distribution of topics.

\printbibliography

\appendix

\newpage
\section{Tables}

\begin{table}[htbp]
\caption{Authors with largest number ot texts}
\begin{tabular}{lrl|llr}
\hline
\multicolumn{3}{c}{All authors} &  &  \multicolumn{2}{c}{Authors of classic prose}  \\ \hline
Author & N. of texts & Comment & & Author & N. of texts \\ 
Unknown Author & 2442 &  &  & Chekhov & 446 \\ \hline
«Tomorrow» & 597 & A weekly publication &  & Maupassant & 390 \\ \hline
Bradbury & 550 &  &  & Gorky & 379 \\ \hline
Bulychev & 540 & \multicolumn{1}{p{4.5cm}|}{Russian sci-fi writer} &  & Tolstoi & 311 \\ \hline
Asimov & 508 &  &  & Grin & 295 \\ \hline
Marina Serova & 464 & \multicolumn{1}{p{4.5cm}|}{A group of mystery fiction writers} &  & P. Neruda & 245 \\ \hline
Anton Chekhov & 446 &  &  & E. A. Poe & 237 \\ \hline
«CompuTerra» & 437 & A weekly publication &  & Nabokov & 227 \\ \hline
Agatha Christie & 433 &  &  & Borges & 224 \\ \hline
Stephen King & 392 &  &  & O. Henry & 217 \\ \hline
Guy de Maupassant & 390 &  &  & Leskov & 215 \\ \hline
Maxim Gorky & 379 &  &  & Mark Twain & 212 \\ \hline
Arthur Conan Doyle & 378 &  &  & Dumas & 185 \\ \hline
Victor Weber & 371 & Translator &  & Kuprin & 183 \\ \hline
Barbara Cartland & 356 &  &  & Kipling & 179 \\ \hline
Robert Sheckley & 356 &  &  & Bunin & 171 \\ \hline
Irina Gurova & 353 & Translator &  & Bulgakov & 165 \\ \hline
Fedor Razzakov & 348 & \multicolumn{1}{p{4.5cm}|}{A biographer of Russian media stars.} &  & Solzhenitsyn & 164 \\ \hline
Stanisław Lem & 341 &  &  & L. Andreev & 153 \\ \hline
Leo Tolstoi & 311 &  &  & Petrushevskaya & 152 \\ \hline
Alexander Grin & 295 & \multicolumn{1}{p{4.5cm}|}{Romantic novels set in a fantasy land} &  & Pushkin & 151 \\ \hline
Vladimir Goldich & 291 & Translator &  & Balzak & 150 \\ \hline
Roger Zelazny & 291 &  &  & Hasek & 147 \\ \hline
Robert E. Howard & 286 &  &  & Shukshin & 146 \\ \hline
Tatiana Pertseva & 275 & Translator &  & Kafka & 144 \\ \hline
\end{tabular}
\label{top_authors}
\end{table}

\end{document}